\documentclass{article} 
\usepackage[final]{colm2026_conference}

\usepackage{microtype}
\usepackage{hyperref}
\usepackage{url}
\usepackage{booktabs}
\usepackage{tabularx}
\usepackage{xcolor}

\usepackage{graphicx} 
\usepackage{natbib}  
\usepackage{caption} 
\usepackage{algorithm}
\usepackage{multirow}
\usepackage{tabularx}
\usepackage{algorithmic}
\usepackage{amsmath}      
\usepackage{amssymb}      
\usepackage{graphicx}     
\usepackage{booktabs}     
\usepackage{microtype}    
\usepackage{xcolor}       
\usepackage{booktabs}
\usepackage{multirow}
\usepackage{graphicx}
\usepackage{caption}
\usepackage{booktabs}     
\usepackage{float}        
\usepackage{xcolor}
\usepackage{makecell}
\usepackage{array} 

\usepackage{newfloat}
\usepackage{listings}
\DeclareCaptionStyle{ruled}{labelfont=normalfont,labelsep=colon,strut=off} 
\floatstyle{ruled}
\newfloat{listing}{tb}{lst}{}
\floatname{listing}{Listing}

\usepackage{latexsym}
\usepackage{subcaption}
\usepackage[T1]{fontenc}

\usepackage[utf8]{inputenc}

\usepackage{microtype}

\usepackage{inconsolata}

\usepackage{graphicx}

\usepackage{lineno}
\usepackage{bm}

\definecolor{darkblue}{rgb}{0, 0, 0.5}
\hypersetup{colorlinks=true, citecolor=darkblue, linkcolor=darkblue, urlcolor=darkblue}

\title{What LLMs Know but Don't Say: \\ Non-generative Prior Extraction for Generalization}

\author{Sara Rezaeimanesh \\
Michigan State University\\
\texttt{rezaeima@msu.edu} \\
\And
Kundan Thind \\
Henry Ford Health \\
\texttt{kthind1@hfhs.org} \\
\And
Farzan Siddiqui \\
Henry Ford Health \\
\texttt{fsiddiq2@hfhs.org} \\
\And
Mohammad Ghassemi \\
Michigan State University \\
\texttt{ghassem3@msu.edu} 
}

\begin{document}

\ifcolmsubmission
\linenumbers
\fi

\maketitle

\begin{abstract}
In domains like medicine and finance, large-scale labeled data is costly and often unavailable, leading to models trained on small datasets that struggle to generalize to real-world populations. Large language models contain extensive knowledge from years of research across these domains. 
We propose LoID (Logit-Informed Distributions), a deterministic method for extracting informative prior distributions for Bayesian logistic and linear regression by directly accessing their token-level predictions. Rather than relying on generated text, we probe the model's confidence in opposing semantic directions (positive vs. negative impact) through carefully constructed sentences. By measuring how consistently the LLM favors one direction across diverse phrasings, we extract the strength and reliability of the model's belief about each feature's influence.
We evaluate LoID on 15 real-world tabular datasets under synthetic out-of-distribution (OOD) settings characterized by covariate shift. We compare our approach against (1) standard weakly informative priors, (2) AutoElicit, a method that prompts LLMs to generate expert-style priors, (3) LLMProcesses, a method that prompts LLMs to generate predictions and (4) an oracle-style upper bound derived from fitting logistic/linear regression on an in-distribution (IID) train set. We assess performance using Area Under the Curve (AUC) and Mean Squared Error (MSE). Across datasets, LoID significantly improves performance over logistic and linear regression models trained on OOD data, recovering \textbf{42\%} of the performance gap relative to the oracle model on average. 
\end{abstract}

\section{Introduction}

Machine learning models used in critical domains like medicine and finance often lack access to large, well-balanced, and comprehensively labeled datasets. Clinical datasets may cover only a single hospital or demographic, while credit-risk tables can be skewed toward historical lending patterns. For example, \citet{purushotham2017benchmarkdeeplearningmodels} benchmarked several deep learning models on clinical prediction tasks—including mortality and length of stay—using the MIMIC-III intensive care electronic health record dataset from a single hospital site. Models trained on small, non-representative samples frequently fail to generalize when confronted with real-world populations that differ from their training distribution \citep{brigato2020closelookdeeplearning,li2024suboptimal,app11020796}. Bayesian methods offer a principled way to encode prior knowledge and mitigate such brittleness, but manually eliciting high-quality priors at scale remains a challenge \citep{mikkola2023priorknowledgeelicitationpast}.

LLMs, trained on vast corpora of scientific literature and domain-specific data \citep{brown2020languagemodelsfewshotlearners}, offer a new source of priors for downstream modeling. Recent efforts have explored LLMs both as zero-shot and few-shot predictors for tabular classification \citep{hegselmann2023tabllmfewshotclassificationtabular,requeima2024LLMProcessesnumericalpredictive}, and as in-context learners capable of producing joint probabilistic predictions \citep{shysheya2025joltjointprobabilisticpredictions}. Other work has investigated the integration of LLMs with classical statistical methods: LLM-FE \citep{abhyankar2025llmfeautomatedfeatureengineering} uses LLMs to guide feature engineering, LLM-Lasso \citep{zhang2025llmlassorobustframeworkdomaininformed} for feature selection, and LLMCorr \citep{zhong2024harnessinglargelanguagemodels} for post-hoc model correction. AutoElicit \citep{capstick2025autoelicitusinglargelanguage} introduced a technique to query LLMs for Gaussian priors, but its reliance on generative sampling introduces run-to-run variability.

The practical benefit of LLM priors depends on how well they generalize under distribution shift. OOD generalization has therefore received increasing attention, as real-world deployment often involves data that differs from the development environment. Meta-learning approaches like TabPFN \citep{hollmann2023tabpfntransformersolvessmall} and Drift-Resilient TabPFN \citep{helli2024driftresilienttabpfnincontextlearning} address OOD robustness through synthetic task generation and causal modeling. Others have proposed logit-based calibration strategies such as UE-NL \citep{huang2023uncertaintyestimationnormalizedlogitsoutofdistribution}, or LLM-guided data augmentation methods like Summary Boosting \citep{manikandan2023languagemodelsweaklearners} and LLMOverTab \citep{ISOMURA2025125852} to cover underrepresented regions of the feature space. However, few works have explored how to extract priors from the internal structure of LLMs (e.g. hidden states, logits, etc.) for tabular tasks under distribution shift.

We address this gap with \textbf{LoID (Logit-Informed Distributions)}, a deterministic framework that elicits informative priors for Bayesian logistic and linear regression directly from an LLM’s token-level logits. Instead of relying on generated completions or examples, LoID probes the LLM with semantically paired sentences highlighting positive and negative feature-target relationships. The logit difference between the relational tokens quantifies directional belief, and the variance across paraphrased prompts yields an uncertainty estimate—together forming Normal priors over model coefficients. These priors integrate into Bayesian workflows, without any fine-tuning or in-context learning.

We evaluate LoID on 14 widely used and publicly available tabular datasets, along with one non-public medical dataset to assess performance on data that does not exist in pre-training corpora. We compare LoID against weakly informative Bayesian inference, LLMProcesses, and AutoElicit. To provide an upper bound, we additionally train logistic and linear regression models on IID training sets, offering insight into performance when the true data distribution is available. Our experiments show that LoID improves OOD generalization across 13/15 tabular datasets compared to $\mathbf{P_{OOD}}$, recovering 42\% of the lost performance on average.

\section{Related work}

Our work draws on two lines of research: (i) using LLMs as a knowledge
source for tabular modeling, and (ii) improving tabular generalization
under distribution shift. We organize each by the mechanism used, since
that is what separates LoID from prior work. We refer the reader to
\citet{fang2024largelanguagemodelsllmstabular} for a broader survey of
LLMs applied to tabular data.

\subsection{LLMs as a knowledge source for tabular modeling}

Existing methods differ in \emph{where} LLM knowledge enters the
pipeline, and fall into three groups:

\textbf{The LLM as the predictor.} TabLLM
\citep{hegselmann2023tabllmfewshotclassificationtabular} serializes rows
into natural language for zero- and few-shot classification;
LLMProcesses \citep{requeima2024LLMProcessesnumericalpredictive}
conditions on numerical data and textual descriptions to produce
probabilistic regression outputs; JoLT
\citep{shysheya2025joltjointprobabilisticpredictions} uses in-context
learning to define joint predictions over tabular data. These methods
require an LLM call per test instance.

\textbf{The LLM as a pipeline component.} A second group keeps a
classical estimator and uses the LLM to shape it: LLM-FE
\citep{abhyankar2025llmfeautomatedfeatureengineering} pairs evolutionary
search with LLM reasoning to construct features, LLM-Lasso
\citep{zhang2025llmlassorobustframeworkdomaininformed} guides feature
selection and regularization, and LLMCorr
\citep{zhong2024harnessinglargelanguagemodels} applies the LLM as a
post-hoc corrector of model predictions. Here the LLM constrains the
hypothesis space, but the fit is driven entirely by the observed
data.

\textbf{The LLM as a prior over parameters.} Prior elicitation has
historically relied on structured interviews with human domain experts,
which is accurate but does not scale
\citep{ohagan2006uncertain, mikkola2023priorknowledgeelicitationpast}.
Recent work automates parts of this process with LLMs:
\citet{babakov2024scalabilitybayesiannetworkstructure} elicit Bayesian
network structure, and AutoElicit
\citep{capstick2025autoelicitusinglargelanguage} prompts an LLM in
multiple expert personas and aggregates the sampled responses into
Gaussian priors over regression coefficients. This is the setting we
work in, and AutoElicit is the closest work to ours.

All three groups access the LLM's knowledge through \emph{generated
text} which makes the extracted quantity
stochastic across runs. LoID instead reads that belief directly off the token-level logits,
which is deterministic and preserves the model's uncertainty as a usable
signal.

\subsection{Generalization under distribution shift}

Approaches to tabular distribution shift differ in what source of
information they use to compensate for unrepresentative training data.

\textbf{Learned inductive biases.} TabPFN
\citep{hollmann2023tabpfntransformersolvessmall} is meta-trained on
millions of synthetic tasks to approximate posterior predictive
distributions in a single forward pass, and Drift-Resilient TabPFN
\citep{helli2024driftresilienttabpfnincontextlearning} extends this with
temporally shifting causal mechanisms to improve robustness under drift.
The prior here is powerful but implicit: it is learned from synthetic
data and cannot be inspected or edited per feature.

\textbf{Robustness and calibration from the training data.}
Distributionally robust methods such as Group DRO \citep{sagawa2020distributionallyrobustneuralnetworks}
reweight training subpopulations to protect worst-group performance,
and calibration methods such as UE-NL
\citep{huang2023uncertaintyestimationnormalizedlogitsoutofdistribution}
regularize logit confidence to improve OOD detection. Both operate
entirely on the observed sample, and so cannot supply information about
regions the training data never covers.

\textbf{Injecting external knowledge.} Closest to our work,
Summary Boosting \citep{manikandan2023languagemodelsweaklearners} uses
LLM-generated feature summaries as weak learners, and LLMOverTab
\citep{ISOMURA2025125852} prompts LLMs to synthesize realistic examples
in underrepresented regions of the feature space. These inject knowledge
at the \emph{data} level. LoID injects it at the \emph{parameter} level which yields priors that are directly
interpretable as beliefs about individual coefficients.

Most prior work on distribution shift has focused on temporally shifted data. In these settings, (i) the test set often does not represent the broader target population and instead constitutes an out-of-distribution subset created by excluding specific groups (e.g., patients older than 55 years or observations collected after 12 p.m.), and (ii) the training set is not explicitly constructed to follow a different distribution or induce degraded downstream performance. In contrast, our goal is to train on an out-of-distribution training set and evaluate whether our method can recover a distribution that more closely reflects the true population distribution. Consequently, the dataset splits used in prior work are generally not suitable for our setting, motivating us to construct our own benchmark and evaluation splits, as described in Section~\ref{subsec:eval3.5}.

\subsection{Choice of baselines}

Our comparisons are chosen to isolate the two decisions that define
LoID: whether to convert LLM knowledge into a prior at all, and how to extract it. AutoElicit
\citep{capstick2025autoelicitusinglargelanguage} shares the first
decision and differs on the second. It targets Gaussian priors for the
same Bayesian models, so the inference procedure, the posterior
approximation, and the training data are held fixed, and any difference
in performance is attributable to the extraction mechanism. LLMProcesses
\citep{requeima2024LLMProcessesnumericalpredictive} differs on the
first: it uses the same LLM and the same OOD train rows, but consumes
them as in-context examples for direct prediction rather than as
evidence to update a prior. It therefore tests whether the detour
through a parametric model is worth taking, or whether the LLM is better
used as the predictor itself. We bracket both with a weakly informative
$\mathcal{N}(0,1)$ prior, which measures the information gain from LLM
knowledge over regularization alone, and with an oracle model fit on IID
data, which bounds the gap any prior could close. We do not compare
against methods that use the LLM to reshape the feature space, as these are complementary to
LoID rather than alternatives — LoID's priors are defined over whatever
features the final model uses, and could be applied downstream of them.

\section{Method}

We propose LoID (Logit-Informed Distributions), a deterministic framework for extracting priors for tabular dataset features from LLMs using token-level logits. Our goal is to derive informative priors for Bayesian logistic and linear regression that generalize robustly in low-data settings.

\subsection{Querying LLM beliefs via logit differences}
\label{subsec:logit-priors}

For each feature $f_j$, we construct paired prompts that express opposing semantic claims about its effect on the target variable. Each prompt is designed such that all relevant context appears before a sentiment-bearing token (e.g., ``positive'' or ``negative''), ensuring that the token's probability considers the entire context for the task, including the name of the feature and the prediction task, $t$ (see examples in \ref{sec:sents}). 

\begin{center}
\texttt{``The impact of $f_j$ on $t$ is ''}
\end{center}

We use the probabilities from the LLM's softmax over its vocabulary to extract the probability of the next token being positive or negative. For the $i$th prompt prefix we define:
\begin{align*}
\mathcal{P}_i^{+} = P(\texttt{<positive\_token>} \mid \text{prefix}), 
\quad \mathcal{P}_i^{-} = P(\texttt{<negative\_token>} \mid \text{prefix})
\end{align*}

We then define the \textbf{logit preference score} as:

\[
\text{logit}_j(p_i) = \log\left(\frac{p_i}{1 - p_i}\right), \quad
p_i = \frac{P_i^+}{P_i^+ + P_i^-}
\]

The logit preference score quantifies the extent to which the model prefers the positive token over the negative token in a given context. While this difference may also reflect other patterns learned during training, our hypothesis is that by constructing sentences that do not grammatically or syntactically favor either token, we reduce the influence of such confounding factors.
Finally, applying the logit transformation maps probabilities from the bounded interval (0, 1) to the unbounded real line, making the resulting scores directly applicable as coefficients in logistic and linear regression models.

\begin{figure*}[t]
  \centering
  \includegraphics[width=1.00\linewidth]{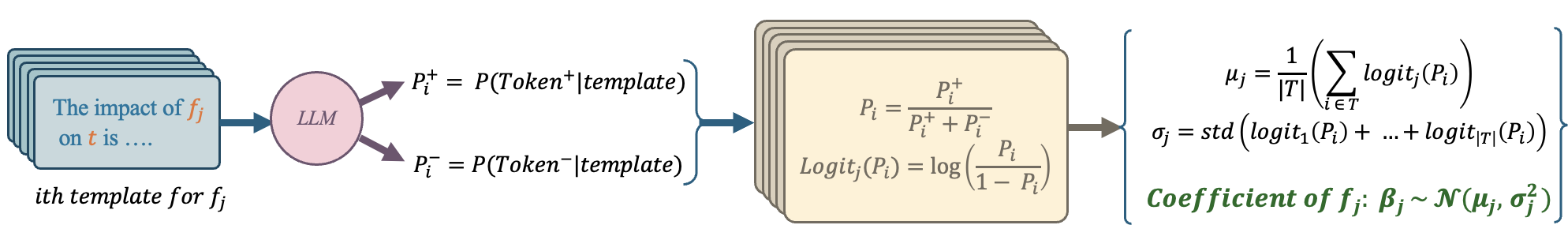}
  \caption{LoID Pipeline: To extract informative priors, LoID uses multiple sentence templates for each feature and computes the probabilities of the tokens "positive" and "negative" as the next token. From these probabilities, we calculate the logit preference score, which quantifies the model’s preference for one token over the other. Using the logit preference scores from all sentences, we compute their mean and standard deviation and use them as informative priors for Bayesian logistic/linear regression.}
  \label{fig:method}
\end{figure*}

\subsection{Constructing priors from logits}
\label{subsec:logit-priors-2}
We model each coefficient $\beta_j$ with a Gaussian prior:
$\beta_j \sim \mathcal{N}(\mu_j, \sigma_j^2)$.
The mean for feature $f_j$ is obtained by averaging the logit
preference score across a set of paraphrased prompts $T$:
\[
\mu_j = \frac{1}{|T|} \sum_{i \in T} \text{logit}_j(p_i)
\]
This value reflects the LLM's normalized preference for a positive
versus negative association with the feature.
We quantify the model's uncertainty about each
feature by the dispersion of its preference scores across the
paraphrased prompts:
\[
\sigma_j = \operatorname{std}\big(\text{logit}_j(p_1), \ldots,
                                  \text{logit}_j(p_{|T|})\big)
\]
where $\text{logit}_j(p_i)$ denotes the logit preference score for the
$i$th prompt and feature $j$. This design assigns higher variance to
features for which the LLM shows inconsistent preferences across
paraphrases, capturing greater uncertainty about their influence. We
evaluated three variance computation methods and found that the
standard deviation of logit preference scores performs best
(Section~\ref{app:var}). Figure~\ref{fig:method} illustrates how LoID extracts informative priors
from model logits.

\subsection{Bayesian inference with LoID priors}

With LoID priors, posterior inference is performed using Markov Chain Monte Carlo (MCMC) sampling via the No-U-Turn Sampler (NUTS) \citep{hoffman2011nouturnsampleradaptivelysetting}.

We run 4 chains with 1,000 tuning iterations and 1,000 sampling iterations each, yielding 4,000 posterior samples total. Predictions are made by marginalizing over these samples:
\begin{equation}
p(y_* \mid x_*, \mathcal{D}_{\text{train}}) \approx \frac{1}{S} \sum_{s=1}^{S} p(y_* \mid x_*, \beta^{(s)}),
\end{equation}
where $\{\beta^{(s)}\}_{s=1}^{S}$ are the posterior samples. For classification, we report the mean predicted probability; for regression, we report the mean prediction.

The NUTS sampler automatically adapts step sizes and uses gradient information to efficiently explore the posterior, with the target acceptance rate set to 0.95 for robust convergence.

\subsection{Evaluation}
\label{subsec:eval3.5}

To evaluate LoID, we construct synthetic OOD splits from real-world datasets. We begin by creating a standard 80--20 random split, where the training portion serves as the in-distribution (IID) training set and the held-out portion serves as the IID test set. To simulate distribution shift, we derive an OOD train set by selecting a subset of the IID training data and excluding specific categories of categorical features or ranges of numerical features. Section \ref{app:ood_splits} explains the synthetic distribution shift for each dataset in detail. 

We verify the presence of distribution shift by comparing the predictive performance of a model trained on the IID training set with that of a model trained on the OOD train set, both evaluated on the same IID test set. A substantial degradation in performance when training on the OOD split indicates that the training data no longer adequately represents the true data distribution.

Given this setup, we then train Bayesian models with LoID priors using the OOD train set and evaluate their performance on the IID test set. Our objective is to assess the extent to which LoID priors can mitigate the performance gap introduced by distribution shift.


\section{Experimental setup}
\subsection{Datasets}

\paragraph{Head and Neck Cancer Recurrence dataset}
We compiled a binary classification dataset for Head and Neck Cancer Recurrence. This dataset includes comprehensive features about patient demographics, tumor characteristics, treatment regimens, and clinical staging, and predicts the likelihood of cancer recurrence following radiotherapy. The dataset encompasses 508 patients with head and neck cancers, capturing clinical variables including comorbidity indices (Charlson score), smoking history (pack-years), HPV status, primary tumor site localization, and detailed radiation therapy parameters such as treatment duration, prescription dose, and the critical time interval between biopsy and treatment initiation. 

This dataset presents a significant class imbalance (390 non-recurrence vs. 118 recurrence cases), reflecting real-world clinical distributions. The dataset is particularly valuable for evaluating model performance under distributional shift, as we created an out-of-distribution test set focusing on elderly patients ($age \geq 75$). This dataset represents a clinically meaningful population that is not present in any LLM pre-training corpora.\footnote{We will release this dataset upon publication.}

\paragraph{Public datasets}
We evaluated LoID on publicly available tabular datasets selected according to the following criteria: (1) features must be semantically meaningful, as LoID relies on language models to infer relationships from feature descriptions; (2) datasets must be tabular, excluding image and time-series data; (3) tasks should exhibit measurable performance degradation under limited-data or distribution-shift settings, providing an opportunity for informative priors to improve predictive performance; and (4) datasets must contain real-world rather than synthetic observations.

Using these criteria, we selected 14 publicly available datasets spanning binary classification and regression tasks: Adult Income \citep{adult_2}, Cervical Cancer \citep{cervical_cancer}, Chronic Kidney Disease \citep{chronic_kidney_disease_336}, Bank Marketing \citep{bank_marketing_222}, Blood Transfusion Service Center \citep{blood_transfusion_service_center_176}, Stroke \citep{fedesoriano2020stroke}, Heart Failure \citep{heart_failure_clinical_records_519}, Diabetes Prediction \citep{mustafatz2023diabetes}, Give Me Some Credit \citep{givemesomecredit2011}, Airfoil Self-Noise \citep{airfoil_self-noise_291}, Bike Sharing \citep{bike_sharing_275}, Combined Cycle Power Plant \citep{combined_cycle_power_plant_294}, Concrete Compressive Strength \citep{concrete_compressive_strength_165}, and Communities and Crime \citep{communities_and_crime_183}.

\paragraph{Preprocessing}
Across all datasets, we apply the same preprocessing pipeline: categorical features are one-hot encoded, missing categorical values are imputed with mode and missing numerical values are imputed with mean. For datasets with more than 15 features, we perform feature selection using Claude-3.5-Sonnet by providing feature descriptions and prompting it to return a list of the 15 most relevant features for the prediction task. The same reduced feature set is used consistently \textbf{across all experimental conditions and baselines} to ensure comparability. We apply z-score normalization to the features for AutoElicit and LoID, but not for LLMProcesses, following the authors’ recommendations.

\paragraph{Development datasets}
We designate the Stroke, Blood Donation, Heart Failure, and Adult Income datasets as development datasets for choosing an LLM and the number of sentence templates in \ref{app:hpo}. The remaining 11 datasets are reserved for testing the generalization performance of the final LoID method.

\subsection{Creating distribution shifts}
Similar to prior work \citep{capstick2025autoelicitusinglargelanguage,helli2024driftresilienttabpfnincontextlearning}, for each dataset, we construct OOD splits as described in Section~\ref{app:ood_splits}. As shown in Table \ref{tab:auc-kl-combined} in the $P_{IID}$ (in-distribution train set performance on in-distribution test set), and $P_{OOD}$ (out-of-distribution train set performance on in-distribution test set) columns, the model trained on the OOD train set noticeably underperforms compared to the model trained on the IID train set. This confirms that generalization is impaired and the OOD train set does not represent the full distribution of the dataset as well as the IID train set.

\begin{table}[t]
\centering
\resizebox{\linewidth}{!}{
\begin{tabular}{@{}lcccccccc@{}}
\toprule
\textbf{Dataset} & $\mathbf{P_{OOD}}$ & $\mathbf{P_{IID}}$ & $\bm{\mathcal{N}}(0,1)$ & \textbf{LoID} & \textbf{LLMProc} & \textbf{AutoEl} & \textbf{Gap (\%)} \\
\midrule

\multicolumn{8}{l}{\textbf{Evaluation Datasets (AUC)}} \\
\midrule
Adult Income & 0.414 & 0.809 & 0.510 & \textbf{0.745} & 0.731 & 0.514 & 84 \\
Stroke & 0.608 & 0.842 & 0.618 & \textbf{0.694} & 0.452 & 0.680 & 37 \\
Heart Failure & 0.746 & 0.849 & 0.742 & 0.753 & 0.608 & \textbf{0.762} & 7 \\
Blood Donation & 0.572 & 0.748 & 0.572 & \textbf{0.663} & 0.516 & 0.577 & 52 \\

\midrule
\multicolumn{8}{l}{\textbf{Additional Classification Benchmarks (AUC)}} \\
\midrule
Head \& Neck Cancer & 0.542 & 0.998 & 0.567 & \textbf{0.696} & 0.540 & 0.609 & 34 \\
Cervical Cancer & 0.440 & 0.870 & 0.448 & \textbf{0.681} & 0.658 & 0.570 & 56 \\
Kidney Disease & 0.822 & 0.993 & 0.715 & \textbf{0.901} & 0.590 & 0.320 & 46 \\
Diabetes & 0.883 & 0.963 & 0.853 & \textbf{0.906} & 0.502 & 0.864 & 29 \\
Bank Marketing & 0.780 & 0.878 & \textbf{0.780} & 0.652 & 0.498 & 0.692 & 0 \\
Good Credit Score & 0.710 & 0.790 & 0.710 & \textbf{0.760} & 0.700 & 0.710 & 63 \\

\midrule
\multicolumn{8}{l}{\textbf{Regression Benchmarks (MSE)}} \\
\midrule
Concrete Strength & 469.20 & 98.14 & 165.72 & \textbf{120.59} & 136.57 & 1974.11 & 94 \\
Communities \& Crime & 0.17 & 0.01 & 0.173 & \textbf{0.07} & 0.15 & 0.30 & 63 \\
Airfoil Noise & 77.16 & 23.69 & 51.11 & \textbf{40.51} & 47.80 & 62.21 & 69 \\
Combined Cycle Power & 514.94 & 21.24 & \textbf{23.55} & 509.92 & 462.89 & 47.14 & 1 \\
Bike Sharing & 77858 & 19030 & 71890 & 78351 & 75650 & \textbf{71405} & 0 \\

\bottomrule
\end{tabular}
}
\caption{Performance comparison across 15 datasets. Gap Closed represents the percentage of the OOD performance gap recovered by our method, LoID. The percentage closed is calculated using: $\frac{max(LoID - P_{OOD},\ 0)}{P_{IID} - P_{OOD}} \times 100$ for classification and $\frac{max(P_{OOD} - LoID,\ 0)}{P_{OOD} - P_{IID}} \times 100$ for regression tasks. The first section contains the four evaluation datasets used in our primary study; the remaining datasets are additional benchmarks used to assess generalization. LoID results are achieved using 10 prompt templates based on the analysis in Section~\ref{sec:sents}. LLMProcesses and AutoElicit use the hyperparameters suggested by their authors as outlined in Section~\ref{app:baseline_hyperparams}. For each baseline method, the results of the best-performing LLM among the tested LLMs is reported (Performance details available in Section~\ref{sec:LLMP}\@). }

\label{tab:auc-kl-combined}
\end{table}

\subsection{LLM-based prior extraction}
\label{sec:experimental-setup}

We use \textbf{Llama-3-8B}, \textbf{Qwen-2.5-14B}, \textbf{Gemma-2-27B}, and \textbf{Gemma-3-27B} to extract logit-based priors as described in Section~\ref{subsec:logit-priors}. We use these models in \textit{zero-shot} mode without any fine-tuning or in-context examples. In our experiments, the queried lexical items are represented as single tokens across models, enabling direct logit-based comparisons.

Each dataset is processed independently. For each feature, we issue 10 prompts to extract token-level log-probabilities for the positive and negative completions. The number of prompts is selected via hyperparameter tuning on the four development datasets (Section~\ref{sec:sents}). The resulting feature-wise prior means and variances are then used to define informative Gaussian priors in Bayesian logistic and linear regression models.

We compare LoID to the following baselines.
\begin{itemize}
    \item $\mathbf{P_{OOD}}$: Logistic/Linear regression model trained on OOD train set. 
    \item \textbf{AutoEl (AutoElicit)}: A Bayesian regression model using LLM-generated expert priors obtained from sampled completions. LoID and AutoElicit differ primarily in their prior elicitation procedures; subsequent Bayesian inference and MCMC sampling are identical. Therefore, performance differences can be attributed to the quality of the extracted priors.
    \item \textbf{LLMProc (LLMProcesses)}: An in-context learning approach that generates predictions by converting tabular features into natural language prompts. Because LLMProcesses does not include a trainable machine learning component, we use OOD train set examples as few-shot samples for the model.
    \item $\mathbf{P_{IID}}$: Performance obtained by training a logistic/linear regression model on the in-distribution (IID) data. This represents the oracle benchmark where the model has full access to the training distribution, serving as a reference point for evaluating OOD generalization.
    \item $\bm{\mathcal{N}}(0,1)$: A weakly informative baseline using standard normal $\mathcal{N}(0,1)$ priors for all feature coefficients. This represents a weakly informative Bayesian approach without incorporating domain knowledge, serving as a reference point to quantify the information gain provided by LLM-derived priors. While a uniform prior would be maximally uninformative, our experiments in Section~\ref{sec:prior_distribution_comparison} show that normal priors are more compatible with the MCMC pipeline and provide a stricter baseline.
    \item \textbf{Additional baselines.} We further compare LoID against correlation-based priors and L1 and L2 regularization (Section~\ref{subsec:addbase}), confirming that its gains do not stem from regularization alone and that frequentist coefficient estimation fails when the training set is OOD.
 \end{itemize}

For AutoElicit and LLMProcesses, we adopt the hyperparameters recommended by the original authors to the extent possible in order to ensure a fair comparison. A comprehensive list of the hyperparameters used to reproduce these baselines is provided in Section \ref{app:baseline_hyperparams}.

\section{Results and discussion}

We evaluate LoID across 15 datasets spanning diverse domains to understand when LoID is able to close the performance gap caused by covariate shift. Our results in Table \ref{tab:auc-kl-combined} reveal a systematic pattern: LoID excels when feature-target relationships are semantically meaningful and generalizable, but struggles when relationships are contextual, behavioral, or dataset-specific.

\subsection{Where LoID helps}

\textbf{Medical and demographic tasks with stable relationships (up to 84\% gap closed).} LoID achieves the strongest gains on datasets where feature-target relationships reflect generalizable biomedical or socioeconomic knowledge:
\begin{itemize}
    \item \textbf{Adult Income} (84\% gap): Age, education, and occupation have well-known effects on income that generalize across demographic shifts
    \item \textbf{Chronic Kidney Disease} (46\% gap): Clinical markers (hemoglobin, albumin) have established diagnostic value
\end{itemize}

In these domains, the LLM's world knowledge captures stable domain relationships that remain valid under distribution shift.

\textbf{Physical systems with known causal structure.} For regression tasks where features have clear physical interpretations (Concrete Strength: 94\% gap, Airfoil Noise: 69\% gap, Communities and Crime: 63\% gap), LoID leverages the LLM's understanding of domain physics and social science to regularize predictions under shift.

\begin{figure}[t]
\centering
\includegraphics[width=1\textwidth]{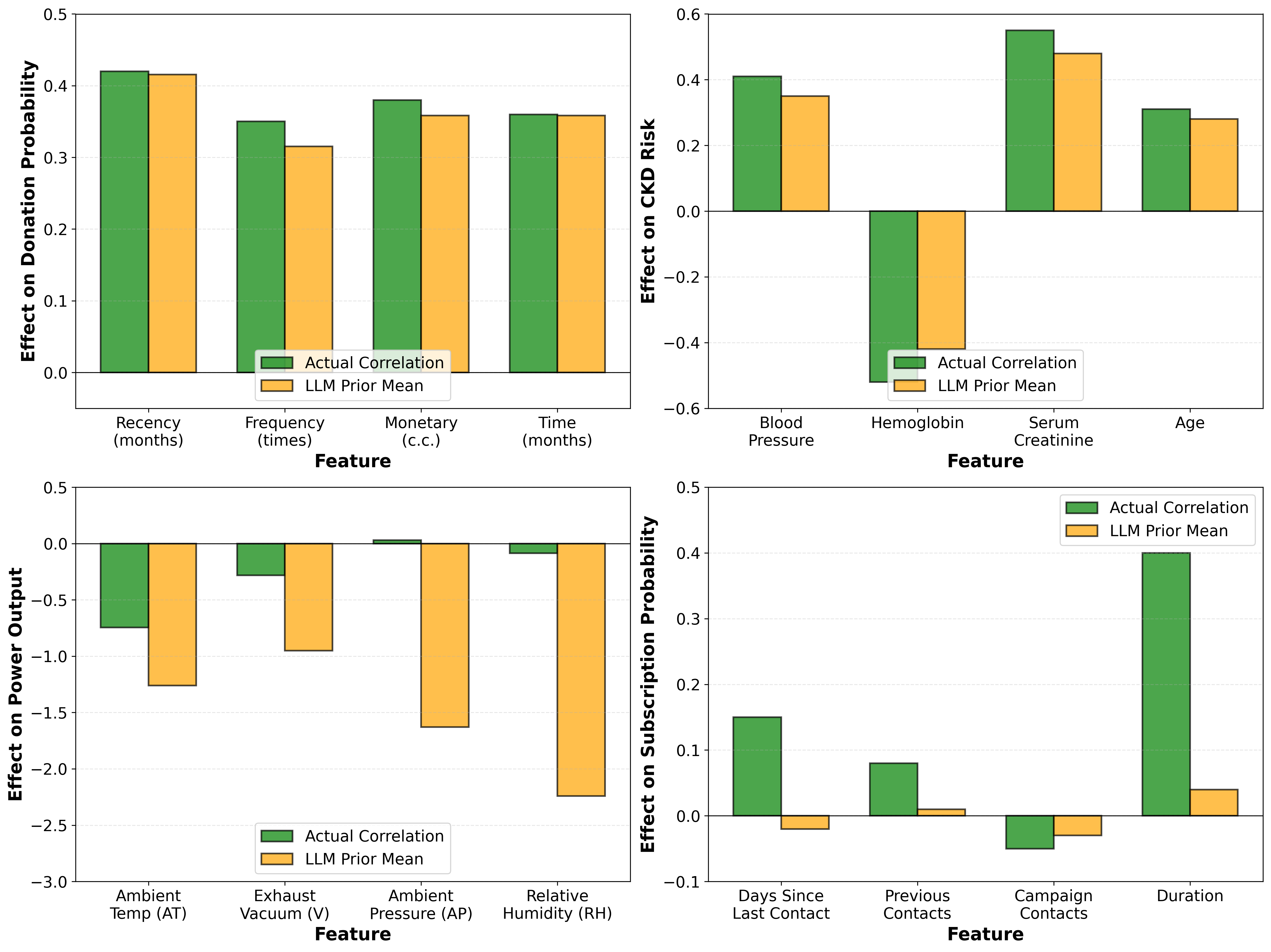}
\caption{LLM prior alignment across success and failure modes. \textbf{Top:} Well-aligned priors for Blood Donation and Chronic Kidney Disease show strong agreement between LLM beliefs and actual correlations. \textbf{Bottom:} Combined Cycle Power exhibits directionally correct but magnitude-misspecified priors, while Bank Marketing shows absent priors (near-zero) for context-dependent features.}
\label{fig:prior_case_studies}
\end{figure}

\subsection{Where LoID fails}
\label{sec:fails}

\textbf{Behavioral and contextual tasks with dataset-specific patterns.} LoID fails on three datasets where relationships are inherently local or behavioral:\footnote{Gap-closed percentages in this section are signed and uncapped to convey both direction and magnitude; the Gap column in Table~\ref{tab:auc-kl-combined} clamps negative values to zero, so these datasets appear as $0$ there.}
\begin{itemize}
    \item \textbf{Bank Marketing} ($-131\%$ gap closed): Customer banking behavior (call duration, previous contacts) varies wildly across campaigns.
    \item \textbf{Bike Sharing} ($-1\%$ gap closed): Demand is driven by local weather, events, and infrastructure, relationships that are city-specific and temporal
    \item \textbf{Combined Cycle Power} ($1\%$ gap closed): Plant efficiency depends on equipment-specific calibrations and operational conditions that LLMs cannot infer from generic turbine descriptions
\end{itemize}

In these cases, the LLM's priors encode \textit{generic relationships} (e.g., ``higher temperature decreases power output'') that may be directionally correct but quantitatively misspecified for the specific plant, city, or campaign. This introduces harmful bias that degrades performance.

\textbf{Understanding failure modes through prior analysis.} Examining the extracted priors reveals two distinct failure modes:

\begin{enumerate}
\item \textbf{Overconfident but wrong priors destroy beneficial regularization.} For Combined Cycle Power, the weakly informative $\mathcal{N}(0,1)$ baseline closes 99.5\% of the OOD gap (MSE: 23.55 vs. P\_OOD: 514.94), demonstrating that simple Bayesian regularization is highly effective. However, LoID closes only 1\% (MSE: 509.92), because the LLM's equipment-agnostic priors (e.g., ``higher ambient temperature decreases turbine efficiency'') override the beneficial regularization with domain knowledge that's directionally plausible but quantitatively misspecified for this specific power plant and severe temperature shift. The LLM knows generic thermodynamics but cannot account for plant-specific calibrations.

\item \textbf{Context-dependent features produce uninformative priors.} For Bank Marketing and Bike Sharing, LLMs have no basis for judging how campaign-specific variables (e.g., ``days since last contact'') or city-specific infrastructure affects outcomes. The extracted priors are either near-zero (indicating maximal uncertainty) or directionally inconsistent, providing no signal over uninformed regularization. Unlike Combined Cycle Power, where the problem is \textit{wrong} priors, here the problem is \textit{absent} priors—the LLM simply lacks relevant knowledge about these context-dependent domains.
\end{enumerate}

Figure~\ref{fig:prior_case_studies} illustrates four representative cases that reveal when and why LLM priors succeed or fail.

\subsection{Comparison to baselines}

\textbf{LoID vs. LLMProcesses.} LLMProcesses achieves near-random performance on 4/10 classification tasks (Stroke, Diabetes, Bank Marketing, Blood Donation). LoID outperforms LLMProcesses on 13/15 datasets. This demonstrates that extracting priors is more effective than direct prediction. The LLM knows the relationships but cannot directly predict outcomes from sparse feature descriptions.

\textbf{LoID vs. AutoElicit.} LLMs tend to produce overconfident priors when explicitly prompted, which can limit the model's ability to adapt to data. As a result, LoID outperforms AutoElicit on 11/15 datasets. For tasks requiring semantic reasoning about feature importance, free-form elicitation may struggle to capture directional relationships as effectively as token-level probability extraction.

\textbf{LoID vs. $\bm{\mathcal{N}}(0,1)$} Comparing LoID to the weakly informative $\mathcal{N}(0,1)$ baseline isolates the value of LLM knowledge. On 12/15 tasks, LoID outperforms $\mathcal{N}(0,1)$, confirming that LLM priors contain useful information. 

\section{Limitations and future work}
While our results demonstrate the effectiveness of LoID, several limitations should be noted, which also point toward directions for future work.
\begin{itemize}
    \item LoID is most useful for datasets with meaningful semantic features, limiting its usage for de-identified datasets.
    \item LoID requires access to token-level logits, limiting compatibility with closed-source APIs. 
\end{itemize}

\section{Conclusion}
We introduced LoID, a method for extracting Bayesian priors from large language models through token-level probability distributions. By probing LLMs for beliefs about feature–target relationships and converting these into structured priors, LoID leverages embedded world knowledge to improve predictive performance under distribution shift.

Across 15 datasets, we observe a consistent pattern: LoID performs strongly when feature–target relationships are semantically stable and generalizable, closing up to 84\% of the performance gap induced by out-of-distribution training in medical and demographic settings, but degrades when these relationships are context-dependent or dataset-specific, as in behavioral tasks such as Bank Marketing. Overall, LoID demonstrates that extracting distributional priors is more robust than direct prediction (compared to LLMProcesses) and more reliable than free-form elicitation (compared to AutoElicit), outperforming LLMProcesses on 13/15 and AutoElicit on 11/15 datasets.


 
\bibliography{colm2026_conference}

@misc{fang2024largelanguagemodelsllmstabular,
      title={Large Language Models(LLMs) on Tabular Data: Prediction, Generation, and Understanding -- A Survey}, 
      author={Xi Fang and Weijie Xu and Fiona Anting Tan and Jiani Zhang and Ziqing Hu and Yanjun Qi and Scott Nickleach and Diego Socolinsky and Srinivasan Sengamedu and Christos Faloutsos},
      year={2024},
      eprint={2402.17944},
      archivePrefix={arXiv},
      primaryClass={cs.CL},
      url={https://arxiv.org/abs/2402.17944}, 
}

@misc{hegselmann2023tabllmfewshotclassificationtabular,
      title={TabLLM: Few-shot Classification of Tabular Data with Large Language Models}, 
      author={Stefan Hegselmann and Alejandro Buendia and Hunter Lang and Monica Agrawal and Xiaoyi Jiang and David Sontag},
      year={2023},
      eprint={2210.10723},
      archivePrefix={arXiv},
      primaryClass={cs.CL},
      url={https://arxiv.org/abs/2210.10723}, 
}

@misc{requeima2024llmprocessesnumericalpredictive,
      title={LLM Processes: Numerical Predictive Distributions Conditioned on Natural Language}, 
      author={James Requeima and John Bronskill and Dami Choi and Richard E. Turner and David Duvenaud},
      year={2024},
      eprint={2405.12856},
      archivePrefix={arXiv},
      primaryClass={stat.ML},
      url={https://arxiv.org/abs/2405.12856}, 
}

@misc{capstick2025autoelicitusinglargelanguage,
      title={AutoElicit: Using Large Language Models for Expert Prior Elicitation in Predictive Modelling}, 
      author={Alexander Capstick and Rahul G. Krishnan and Payam Barnaghi},
      year={2025},
      eprint={2411.17284},
      archivePrefix={arXiv},
      primaryClass={cs.LG},
      url={https://arxiv.org/abs/2411.17284}, 
}

@misc{abhyankar2025llmfeautomatedfeatureengineering,
      title={LLM-FE: Automated Feature Engineering for Tabular Data with LLMs as Evolutionary Optimizers}, 
      author={Nikhil Abhyankar and Parshin Shojaee and Chandan K. Reddy},
      year={2025},
      eprint={2503.14434},
      archivePrefix={arXiv},
      primaryClass={cs.LG},
      url={https://arxiv.org/abs/2503.14434}, 
}

@misc{zhong2024harnessinglargelanguagemodels,
      title={Harnessing Large Language Models as Post-hoc Correctors}, 
      author={Zhiqiang Zhong and Kuangyu Zhou and Davide Mottin},
      year={2024},
      eprint={2402.13414},
      archivePrefix={arXiv},
      primaryClass={cs.LG},
      url={https://arxiv.org/abs/2402.13414}, 
}

@misc{babakov2024scalabilitybayesiannetworkstructure,
      title={Scalability of Bayesian Network Structure Elicitation with Large Language Models: a Novel Methodology and Comparative Analysis}, 
      author={Nikolay Babakov and Ehud Reiter and Alberto Bugarin},
      year={2024},
      eprint={2407.09311},
      archivePrefix={arXiv},
      primaryClass={cs.CL},
      url={https://arxiv.org/abs/2407.09311}, 
}

@misc{zhang2025llmlassorobustframeworkdomaininformed,
      title={LLM-Lasso: A Robust Framework for Domain-Informed Feature Selection and Regularization}, 
      author={Erica Zhang and Ryunosuke Goto and Naomi Sagan and Jurik Mutter and Nick Phillips and Ash Alizadeh and Kangwook Lee and Jose Blanchet and Mert Pilanci and Robert Tibshirani},
      year={2025},
      eprint={2502.10648},
      archivePrefix={arXiv},
      primaryClass={cs.LG},
      url={https://arxiv.org/abs/2502.10648}, 
}

@misc{brigato2020closelookdeeplearning,
      title={A Close Look at Deep Learning with Small Data}, 
      author={L. Brigato and L. Iocchi},
      year={2020},
      eprint={2003.12843},
      archivePrefix={arXiv},
      primaryClass={cs.LG},
      url={https://arxiv.org/abs/2003.12843}, 
}

@article{li2024suboptimal,
  author    = {Li, Guoxin and Li, Chang and Wang, Chun and Wang, Zhenyu},
  title     = {Suboptimal capability of individual machine learning algorithms in modeling small-scale imbalanced clinical data of local hospital},
  journal   = {PLOS ONE},
  volume    = {19},
  number    = {2},
  pages     = {e0298328},
  year      = {2024},
  month     = feb,
  doi       = {10.1371/journal.pone.0298328},
  pmid      = {38394317},
  pmcid     = {PMC10890755},
  publisher = {Public Library of Science}
}

@Article{app11020796,
AUTHOR = {Althnian, Alhanoof and AlSaeed, Duaa and Al-Baity, Heyam and Samha, Amani and Dris, Alanoud Bin and Alzakari, Najla and Abou Elwafa, Afnan and Kurdi, Heba},
TITLE = {Impact of Dataset Size on Classification Performance: An Empirical Evaluation in the Medical Domain},
JOURNAL = {Applied Sciences},
VOLUME = {11},
YEAR = {2021},
NUMBER = {2},
ARTICLE-NUMBER = {796},
URL = {https://www.mdpi.com/2076-3417/11/2/796},
ISSN = {2076-3417},
DOI = {10.3390/app11020796}
}

@book{ohagan2006uncertain,
  title     = {Uncertain Judgements: Eliciting Experts' Probabilities},
  author    = {O'Hagan, Anthony and Buck, Christopher E. and Daneshkhah, Alireza and Eiser, J. Richard and Garthwaite, Paul H. and Jenkinson, David J. and Oakley, Jeremy E. and Rakow, Tim},
  year      = {2006},
  publisher = {John Wiley \& Sons},
  address   = {Chichester, UK},
  isbn      = {978-0-470-02999-5}
}

@misc{brown2020languagemodelsfewshotlearners,
      title={Language Models are Few-Shot Learners}, 
      author={Tom B. Brown and Benjamin Mann and Nick Ryder and Melanie Subbiah and Jared Kaplan and Prafulla Dhariwal and Arvind Neelakantan and Pranav Shyam and Girish Sastry and Amanda Askell and Sandhini Agarwal and Ariel Herbert-Voss and Gretchen Krueger and Tom Henighan and Rewon Child and Aditya Ramesh and Daniel M. Ziegler and Jeffrey Wu and Clemens Winter and Christopher Hesse and Mark Chen and Eric Sigler and Mateusz Litwin and Scott Gray and Benjamin Chess and Jack Clark and Christopher Berner and Sam McCandlish and Alec Radford and Ilya Sutskever and Dario Amodei},
      year={2020},
      eprint={2005.14165},
      archivePrefix={arXiv},
      primaryClass={cs.CL},
      url={https://arxiv.org/abs/2005.14165}, 
}

@misc{shysheya2025joltjointprobabilisticpredictions,
      title={JoLT: Joint Probabilistic Predictions on Tabular Data Using LLMs}, 
      author={Aliaksandra Shysheya and John Bronskill and James Requeima and Shoaib Ahmed Siddiqui and Javier Gonzalez and David Duvenaud and Richard E. Turner},
      year={2025},
      eprint={2502.11877},
      archivePrefix={arXiv},
      primaryClass={stat.ML},
      url={https://arxiv.org/abs/2502.11877}, 
}

@misc{hollmann2023tabpfntransformersolvessmall,
      title={TabPFN: A Transformer That Solves Small Tabular Classification Problems in a Second}, 
      author={Noah Hollmann and Samuel Müller and Katharina Eggensperger and Frank Hutter},
      year={2023},
      eprint={2207.01848},
      archivePrefix={arXiv},
      primaryClass={cs.LG},
      url={https://arxiv.org/abs/2207.01848}, 
}

@misc{helli2024driftresilienttabpfnincontextlearning,
      title={Drift-Resilient TabPFN: In-Context Learning Temporal Distribution Shifts on Tabular Data}, 
      author={Kai Helli and David Schnurr and Noah Hollmann and Samuel Müller and Frank Hutter},
      year={2024},
      eprint={2411.10634},
      archivePrefix={arXiv},
      primaryClass={cs.LG},
      url={https://arxiv.org/abs/2411.10634}, 
}

@misc{huang2023uncertaintyestimationnormalizedlogitsoutofdistribution,
      title={Uncertainty-Estimation with Normalized Logits for Out-of-Distribution Detection}, 
      author={Mouxiao Huang and Yu Qiao},
      year={2023},
      eprint={2302.07608},
      archivePrefix={arXiv},
      primaryClass={cs.LG},
      url={https://arxiv.org/abs/2302.07608}, 
}

@misc{manikandan2023languagemodelsweaklearners,
      title={Language models are weak learners}, 
      author={Hariharan Manikandan and Yiding Jiang and J Zico Kolter},
      year={2023},
      eprint={2306.14101},
      archivePrefix={arXiv},
      primaryClass={cs.LG},
      url={https://arxiv.org/abs/2306.14101}, 
}

@article{ISOMURA2025125852,
title = {LLMOverTab: Tabular data augmentation with language model-driven oversampling},
journal = {Expert Systems with Applications},
volume = {264},
pages = {125852},
year = {2025},
issn = {0957-4174},
doi = {https://doi.org/10.1016/j.eswa.2024.125852},
url = {https://www.sciencedirect.com/science/article/pii/S0957417424027192},
author = {Tokimasa Isomura and Ryotaro Shimizu and Masayuki Goto},
}

@misc{purushotham2017benchmarkdeeplearningmodels,
      title={Benchmark of Deep Learning Models on Large Healthcare MIMIC Datasets}, 
      author={Sanjay Purushotham and Chuizheng Meng and Zhengping Che and Yan Liu},
      year={2017},
      eprint={1710.08531},
      archivePrefix={arXiv},
      primaryClass={cs.LG},
      url={https://arxiv.org/abs/1710.08531}, 
}

@misc{givemesomecredit2011,
  title        = {{Give Me Some Credit}},
  author       = {Kaggle},
  year         = {2011},
  howpublished = {Kaggle Competition},
  url          = {https://www.kaggle.com/c/GiveMeSomeCredit}
}

@misc{adult_2,
  author       = {Becker, Barry and Kohavi, Ronny},
  title        = {{Adult}},
  year         = {1996},
  howpublished = {UCI Machine Learning Repository},
  note         = {{DOI}: https://doi.org/10.24432/C5XW20}
}

@misc{mustafatz2023diabetes,
  author = {Mustafatz},
  title = {Diabetes prediction dataset},
  year = {2023},
  publisher = {Kaggle},
  howpublished = {\url{https://www.kaggle.com/datasets/iammustafatz/diabetes-prediction-dataset}},
  note = {Accessed: 2026-03-26}
}

@misc{bike_sharing_275,
  author       = {Fanaee-T, Hadi},
  title        = {{Bike Sharing}},
  year         = {2013},
  howpublished = {UCI Machine Learning Repository},
  note         = {{DOI}: https://doi.org/10.24432/C5W894}
}

@misc{combined_cycle_power_plant_294,
  author       = {Tfekci, Pnar and Kaya, Heysem},
  title        = {{Combined Cycle Power Plant}},
  year         = {2014},
  howpublished = {UCI Machine Learning Repository},
  note         = {{DOI}: https://doi.org/10.24432/C5002N}
}

@misc{concrete_compressive_strength_165,
  author       = {Yeh, I-Cheng},
  title        = {{Concrete Compressive Strength}},
  year         = {1998},
  howpublished = {UCI Machine Learning Repository},
  note         = {{DOI}: https://doi.org/10.24432/C5PK67}
}

@misc{communities_and_crime_183,
  author       = {Redmond, Michael},
  title        = {{Communities and Crime}},
  year         = {2002},
  howpublished = {UCI Machine Learning Repository},
  note         = {{DOI}: https://doi.org/10.24432/C53W3X}
}

@misc{fedesoriano2020stroke,
  author = {Fedesoriano},
  title = {Stroke Prediction Dataset},
  year = {2020},
  publisher = {Kaggle},
  howpublished = {\url{https://www.kaggle.com/datasets/fedesoriano/stroke-prediction-dataset}},
  note = {Accessed: 2026-03-26}
}

@article{kim2020cannabis_hnc,
  author  = {Kim, J. and Hua, G. and Zhang, H. and Chan, T. J. and Xie, M. and Levin, M. and Farrokhyar, F. and Archibald, S. D. and Jackson, B. and Young, J. E. and Gupta, M.},
  title   = {Rate of Second Primary Head and Neck Cancer With Cannabis Use},
  journal = {Cureus},
  year    = {2020},
  volume  = {12},
  number  = {11},
  pages   = {e11483},
  doi     = {10.7759/cureus.11483},
  pmid    = {33329979},
  pmcid   = {PMC7735528}
}

@misc{mikkola2023priorknowledgeelicitationpast,
      title={Prior knowledge elicitation: The past, present, and future}, 
      author={Petrus Mikkola and Osvaldo A. Martin and Suyog Chandramouli and Marcelo Hartmann and Oriol Abril Pla and Owen Thomas and Henri Pesonen and Jukka Corander and Aki Vehtari and Samuel Kaski and Paul-Christian Bürkner and Arto Klami},
      year={2023},
      eprint={2112.01380},
      archivePrefix={arXiv},
      primaryClass={stat.ME},
      url={https://arxiv.org/abs/2112.01380}, 
}

@misc{cervical_cancer,
  author       = {Fernandes, Kelwin and Cardoso, Jaime and Fernandes, Jessica},
  title        = {{Cervical Cancer (Risk Factors)}},
  year         = {2017},
  howpublished = {UCI Machine Learning Repository},
  note         = {{DOI}: https://doi.org/10.24432/C5Z310}
}

@misc{chronic_kidney_disease_336,
  author       = {Rubini, L. and Soundarapandian, P. and Eswaran, P.},
  title        = {{Chronic Kidney Disease}},
  year         = {2015},
  howpublished = {UCI Machine Learning Repository},
  note         = {{DOI}: https://doi.org/10.24432/C5G020}
}

@misc{bank_marketing_222,
  author       = {Moro, S. and Rita, P. and Cortez, P.},
  title        = {{Bank Marketing}},
  year         = {2014},
  howpublished = {UCI Machine Learning Repository},
  note         = {{DOI}: https://doi.org/10.24432/C5K306}
}

@misc{blood_transfusion_service_center_176,
  author       = {Yeh, I-Cheng},
  title        = {{Blood Transfusion Service Center}},
  year         = {2008},
  howpublished = {UCI Machine Learning Repository},
  note         = {{DOI}: https://doi.org/10.24432/C5GS39}
}

@misc{heart_failure_clinical_records_519,
  author       = {{UCI Machine Learning Repository}},
  title        = {{Heart Failure Clinical Records}},
  year         = {2020},
  howpublished = {UCI Machine Learning Repository},
  note         = {{DOI}: https://doi.org/10.24432/C5Z89R}
}

@misc{airfoil_self-noise_291,
  author       = {Brooks, Thomas and Pope, D. and Marcolini, Michael},
  title        = {{Airfoil Self-Noise}},
  year         = {1989},
  howpublished = {UCI Machine Learning Repository},
  note         = {{DOI}: https://doi.org/10.24432/C5VW2C}
}

@misc{sagawa2020distributionallyrobustneuralnetworks,
      title={Distributionally Robust Neural Networks for Group Shifts: On the Importance of Regularization for Worst-Case Generalization}, 
      author={Shiori Sagawa and Pang Wei Koh and Tatsunori B. Hashimoto and Percy Liang},
      year={2020},
      eprint={1911.08731},
      archivePrefix={arXiv},
      primaryClass={cs.LG},
      url={https://arxiv.org/abs/1911.08731}, 
}

@misc{hoffman2011nouturnsampleradaptivelysetting,
      title={The No-U-Turn Sampler: Adaptively Setting Path Lengths in Hamiltonian Monte Carlo}, 
      author={Matthew D. Hoffman and Andrew Gelman},
      year={2011},
      eprint={1111.4246},
      archivePrefix={arXiv},
      primaryClass={stat.CO},
      url={https://arxiv.org/abs/1111.4246}, 
}
\bibliographystyle{colm2026_conference}

\newpage

\appendix

\section{Appendix}
\label{sec:appendix}

\subsection{Baseline hyperparameters}
\label{app:baseline_hyperparams}

This section details the hyperparameters used for the baseline methods compared in our experiments. Table~\ref{tab:baseline_hyperparams} provides a comprehensive overview of all configuration settings.

\begin{table*}[h]
\centering
\small
\begin{tabularx}{\linewidth}{l *{3}{>{\raggedright\arraybackslash}X}}
\toprule
\textbf{Category} & \textbf{LLMProcesses} & \textbf{AutoElicit} & \textbf{LoID} \\
\midrule
\multicolumn{4}{l}{\textit{Model Configuration}} \\
Model & Gemma-2-27B-it & Gemma-2-27B-it & Gemma-2-27B-it \\
Temperature & 1.0 & 1.0 & N/A (no sampling) \\
Top-p & 1.0 & 0.9 & N/A \\
Max length & 20 tokens & 8192 tokens & N/A \\
\midrule
\multicolumn{4}{l}{\textit{Prior/Prediction Extraction}} \\
Method & Monte Carlo sampling & Multi-expert prompting & Token probability \\
Samples per point & 50 & 500 (prior predictive) & 10 templates \\
Aggregation & Median & MCMC mixture & Mean and Variance of log-probs \\
Templates/Roles & N/A & 100 (10×10 role pairs) & 10 sentence templates \\
\midrule
\multicolumn{4}{l}{\textit{Data Configuration}} \\
Training samples & 50 (30 for HNC) & 25 & N/A (no in-context) \\
Prompt ordering & Distance-based & N/A & N/A \\
\midrule
\multicolumn{4}{l}{\textit{Reproducibility}} \\
Random seed & N/A & 42 & 42 \\
Repetitions & 5 & 5 & 5 \\
\bottomrule
\end{tabularx}
\caption{Hyperparameter configurations for AutoElicit, LLMProcesses, and LoID. AutoElicit and LLMProcesses use the hyperparameters suggested by their authors.}
\label{tab:baseline_hyperparams}
\end{table*}

\subsection{Hyperparameter optimization}
\label{app:hpo}

We used four datasets---Stroke, Blood Donation, Heart Failure, and Adult Income---as development datasets to tune hyperparameters based on AUC and MSE.

\subsubsection{Number of prompt templates}
\label{sec:sents}
We tested our method with 2--20 prompt templates and 
analyzed how the number of templates affects LoID's performance. The following are some examples of our prompt templates:
\begin{enumerate}
    \item The impact of {} on {} is \ldots
    \item The relationship between {} and {} is \ldots
    \item The role of {} in {} is \ldots
    \item When considering {}, the effect on {} is \ldots
    \item The correlation between {} and {} is \ldots
\end{enumerate}

\begin{figure*}[!h]
    \centering
    \includegraphics[width=1\linewidth]{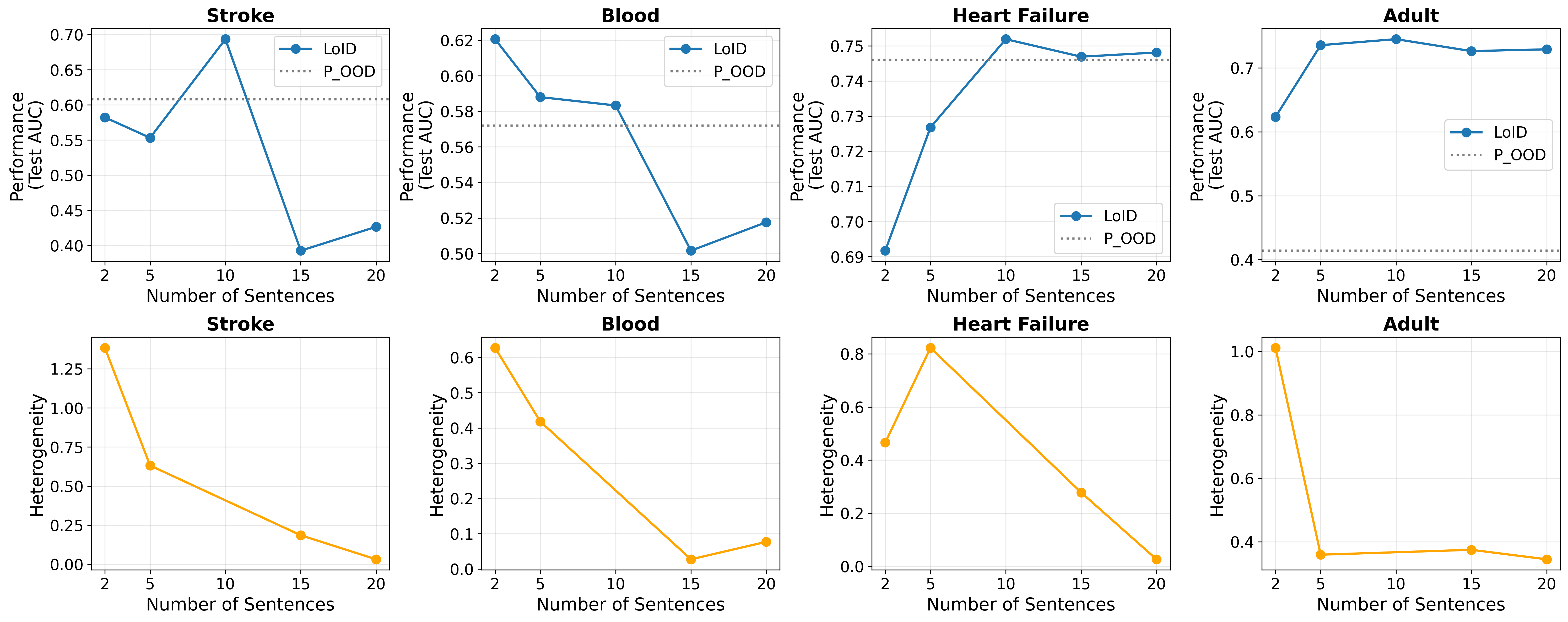}
    \caption{Template count sensitivity analysis across four datasets. Top two rows: Performance (Test AUC) versus number of sentence templates, comparing LoID to out-of-distribution baseline (P\_OOD). Performance peaks at 10 templates for most datasets, with degradation at 15+ templates suggesting overfitting to misspecified priors. Bottom two rows: Prior heterogeneity measured by coefficient of variation (CV) of extracted prior standard deviations. Higher CV indicates greater variance in uncertainty across features. Heterogeneity decreases with more templates as averaging stabilizes estimates, but excessive templates (15+) harm performance without further reducing uncertainty.}
    \label{fig:ablation}
\end{figure*}

Figure~\ref{fig:ablation} reveals the optimal template count for prior extraction. Performance peaks at 10 templates across most datasets, balancing signal aggregation against overfitting. Beyond 10 templates, performance degrades sharply (e.g., Stroke drops from 0.69 to 0.39 at 15 templates), suggesting that excessive prompting amplifies misspecified priors rather than refining them. While coefficient of variation decreases with more templates (indicating more stable uncertainty estimates), this stability comes at the cost of performance when templates exceed 10. Notably, heterogeneity reduction plateaus after 10 templates, yet performance continues to degrade, demonstrating that template averaging helps only when the underlying priors are well-specified. For datasets where LLM knowledge is weak (e.g., Blood Donation), even 2 templates outperform 15+, as additional prompting merely reinforces uninformative/incorrect beliefs. This pattern supports our recommendation of 10 templates as the optimal trade-off between variance reduction and avoiding overconfident misspecification.

\subsubsection{Model selection}
\label{sec:LLMP}

We compared the performances of LLMs-- Llama-3-8B, Qwen-2.5-14B, Gemma-2-27B-Instruct, and Gemma-3-27B-Instruct, in terms of AUC. Table~\ref{tab:llm_comparison} demonstrates that LoID consistently benefits from stronger prior extraction across different LLM backbones, outperforming both AutoElicit and LLMProcesses overall. Among the evaluated models, Gemma-2-27B provides the strongest performance for LoID and matches the best AutoElicit backbone (tied with Llama-3-8B). In contrast, LLMProcesses achieves its best performance with Llama-3-8B but remains substantially below LoID, indicating that directly using LLM predictions through in-context learning is less effective than incorporating LLM-derived priors into a Bayesian model. These results highlight that the quality of the elicited priors, rather than the LLM backbone alone, is a key factor determining downstream performance. 

Interestingly, Gemma-3-27B performed worse on average than Gemma-2-27B despite being a newer model. Inspection of the extracted priors suggests that Gemma-3 frequently assigns higher confidence to its predictions than Gemma-2. While stronger confidence can improve performance when the underlying beliefs are accurate, it also increases the influence of incorrect beliefs on the posterior. As a result, overconfident but inaccurate priors may lead to poorer downstream performance, highlighting the importance of prior calibration in addition to prior accuracy.

To ensure fairness, we use the results from the best-performing LLM for each method in Table \ref{tab:auc-kl-combined}.

\begin{table*}[t]
\centering
\small
\begin{tabular*}{\textwidth}{@{\extracolsep{\fill}}llcccc@{}}
\toprule
\textbf{Method} & \textbf{Dataset} & \textbf{Gemma-2-27B} & \textbf{Gemma-3-27B} & \textbf{Llama-3-8B} & \textbf{Qwen-2.5-14B} \\
\midrule
\multirow{4}{*}{LoID}
& Adult Income & 0.745 & 0.706 & 0.667 & 0.494 \\
& Stroke & 0.694 & 0.633 & 0.679 & 0.604 \\
& Heart Failure & 0.753 & 0.743 & 0.748 & 0.749 \\
& Blood Donation & 0.663 & 0.588 & 0.590 & 0.568 \\
\cmidrule{2-6}
& \textit{Average} & \textit{0.714} & \textit{0.668} & \textit{0.671} & \textit{0.604} \\
\midrule
\multirow{4}{*}{AutoElicit}
& Adult Income & 0.514 & 0.530 & 0.502 & 0.510 \\
& Stroke & 0.680 & 0.665 & 0.549 & 0.670 \\
& Heart Failure & 0.762 & 0.697 & 0.747 & 0.748 \\
& Blood Donation & 0.577 & 0.554 & 0.734 & 0.556 \\
\cmidrule{2-6}
& \textit{Average} & \textit{0.633} & \textit{0.612} & \textit{0.633} & \textit{0.621} \\
\midrule
\multirow{4}{*}{LLMProcesses}
& Adult Income & 0.686 & 0.690 & 0.731 & 0.500 \\
& Stroke & 0.495 & 0.483 & 0.452 & 0.492 \\
& Heart Failure & 0.532 & 0.530 & 0.608 & 0.563 \\
& Blood Donation & 0.500 & 0.502 & 0.516 & 0.494 \\
\cmidrule{2-6}
& \textit{Average} & \textit{0.553} & \textit{0.551} & \textit{0.577} & \textit{0.512} \\
\bottomrule
\end{tabular*}
\caption{Performance comparison of LoID, AutoElicit, and LLMProcesses across different LLMs and datasets.}
\label{tab:llm_comparison}
\end{table*}

\subsubsection{Variance estimation formula}
\label{app:var}
We compared two approaches for estimating prior uncertainty using Gemma-2-27B: variance of normalized logit probabilities and entropy-based uncertainty. The logit variance approach consistently achieved better performance across datasets, improving average AUC from 0.661 to 0.714 compared to entropy-based uncertainty. The largest improvements were observed on Adult Income and Blood Donation datasets, where entropy-based uncertainty resulted in substantially lower performance, with AUC dropping from 0.745 to 0.612 on Adult Income and from 0.663 to 0.598 on Blood Donation. These results suggest that measuring variability in the LLM's estimated effect direction across paraphrased prompts provides a more informative uncertainty estimate than entropy over the token probability distribution.

\paragraph{Method 1: variance of normalized logit probabilities} In this method, $p_j$ reflects the model’s preference between logits. Applying a log transform maps probabilities from $[0,1]$ to the real line, aligning with the unbounded nature of regression coefficients. Variance across paraphrased inputs captures inconsistency in LLM responses, indicating greater uncertainty.

\[
\sigma_j = \text{std}\big(\text{logit}_j(p_1), \ldots, \text{logit}_j(p_{|T|})\big)
\]

where $$\text{logit}_j(p_i) = \log\left(\frac{p_i}{1 - p_i}\right)$$

and $$p_i = \frac{P_i^+}{P_i^+ + P_i^-}
$$

\paragraph{Method 2: entropy-based uncertainty}

This method uses Shannon entropy to measure the uncertainty in the LLM's token probability distribution. Higher entropy (closer to $\log 2 \approx 0.693$) indicates the LLM is uncertain between the positive and negative tokens, reflecting ambiguity about the feature's effect direction. Lower entropy indicates confident predictions. 

\[
\sigma_j = \frac{1}{N_{\text{sent}}}\sum_{i=1}^{N_{\text{sent}}} H({P_i^-}, P_i^+)
\]

\subsubsection{$\mathcal{N}$(0, 1) vs. $\mathcal{U}$(-1, 1)}
\label{sec:prior_distribution_comparison}

To understand the impact of prior distribution choice, we compare two configurations: (1) weakly informative Normal priors $\mathcal{N}(0, 1)$, and (2) uninformative Uniform priors $\mathcal{U}(-1, 1)$.

Table \ref{tab:uniform} reveals that while Uniform priors ostensibly provide ``maximum uninformativeness'' by assigning equal probability across a range, they fail as regularizers for regression coefficients because:

\begin{enumerate}
    \item \textbf{Lack of soft regularization:} Uniform distributions do not encode the reasonable prior belief that smaller coefficient magnitudes are more plausible than larger ones. This absence of gradient-based regularization toward zero can lead to poor MCMC sampling behavior and overfitting.
    
    \item \textbf{Incompatibility with MCMC:} The flat density of Uniform priors provides no gradient information to guide sampling algorithms, resulting in inefficient exploration and convergence issues.

\end{enumerate}

\begin{table*}[h]
\centering
\begin{tabularx}{\linewidth}{l *{3}{>{\centering\arraybackslash}X}}
\toprule
Dataset & ${\mathcal{N}}(0,1)$ & ${\mathcal{U}}(-1,1)$ & $\Delta$ \\
\midrule
Adult Income         & 0.510 & 0.526 & +0.016 \\
Stroke               & 0.618 & 0.392 & -0.226 \\
Heart Failure        & 0.742 & 0.714 & -0.028 \\
Blood Donation       & 0.572 & 0.564 & -0.008 \\
\midrule
\emph{Average} & 0.611 & 0.549 & -0.062 \\
\bottomrule
\end{tabularx}
\caption{Comparison of Bayesian logistic/linear regression with Normal and Uniform priors. Normal priors consistently outperform Uniform priors, likely due to their implicit regularization and improved compatibility with MCMC inference. Consequently, Bayesian logistic/linear regression with Normal priors is adopted as the baseline throughout the paper, providing a stronger point of comparison.}
\label{tab:uniform}
\end{table*}

\subsection{Additional experiments}

\subsubsection{Feature synonyms}
We evaluated the impact of feature synonyms on model performance to determine whether changing feature names could introduce bias based on word selection. To do this, we created three sets of synonyms for the features in our evaluation datasets. Table~\ref{tab:feature_variant_examples} presents examples of these features and their corresponding synonyms. We test three variants: (1) concise, (2) verbose, and (3) clinical feature descriptions. Examples of these variants are provided in Table \ref{tab:feature_variant_examples}.

\paragraph{Results} Table~\ref{tab:adon} shows the effect of different grammatical styles for describing features. We find that Feature description style shows modest and inconsistent effects, which indicates that while LLMs are not entirely robust to paraphrasing, the sensitivity is unpredictable and relatively small.

\begin{table}[h]
\centering

\small
\begin{tabular}{p{2.5cm} p{11cm}}
\toprule
\textbf{Variant} & \textbf{Example Prompt} \\
\midrule
\textbf{Baseline} 
 & The impact of \emph{having hypertension} on risk of stroke is \ldots  \\
 & The impact of \emph{patient's age} on risk of stroke is \ldots \\
\midrule
\textbf{Concise}
& The impact of \emph{hypertension} on risk of stroke is \ldots  \\
 & The impact of \emph{age} on risk of stroke is \ldots \\
\midrule
\textbf{Verbose} 
& The impact of the \emph{patient having a diagnosis of hypertension} on risk of stroke is \ldots \\
 & The impact of \emph{the number of years the patient has lived} on risk of stroke is \ldots\\
\midrule
\textbf{Clinical}
 & The impact of patient \emph{presenting with hypertension} on risk of stroke is \ldots \\
 & The impact of patient \emph{demonstrating advanced age} on risk of stroke is \ldots \\
\bottomrule
\end{tabular}
\caption{Example prompts for feature description variants. All variants use the same template structure but differ in how features are described. The results reported in Table \ref{tab:auc-kl-combined} are achieved using the baseline feature descriptions.}
\label{tab:feature_variant_examples}
\end{table}

\subsubsection{Contextual preambles}
We test whether providing domain-knowledge context before feature-effect questions improves prior extraction. For each feature, we prepend an explanatory preamble describing the causal mechanism, extracted from relevant literature (e.g., ``Stroke risk increases significantly with age due to arterial stiffening and accumulated vascular damage. Based on this medical knowledge,'') before asking the standard question. Table~\ref{tab:preamble_examples} shows example prompts with and without contextual preambles.

Table~\ref{tab:adon} shows the effect of adding domain-knowledge context before feature-effect questions.

\begin{table*}[t]
\centering
\small
\begin{tabularx}{\linewidth}{l *{7}{>{\centering\arraybackslash}X}}
\toprule
& \multicolumn{4}{c}{\textbf{Feature Description Variants}} & \multicolumn{3}{c}{\textbf{Contextual Preambles}} \\
\cmidrule(lr){2-5} \cmidrule(lr){6-8}
\textbf{Dataset} & \textbf{Variant 1} & \textbf{Variant 2} & \textbf{Variant 3} & \textbf{Range} & \textbf{No Preamble} & \textbf{With Preamble} & \textbf{$\Delta$} \\
 & \textbf{(Concise)} & \textbf{(Verbose)} & \textbf{(Clinical)} & & \textbf{(Baseline)} & & \\
\midrule
Adult Income     & 0.753 & 0.636 & 0.767 & 0.131 & 0.745 & 0.771 & +0.026 \\
Stroke           & 0.685 & 0.703 & 0.704 & 0.019 & 0.694 & 0.707 & +0.013 \\
Heart Failure    & 0.713 & 0.736 & 0.764 & 0.051 & 0.753 & 0.638 & -0.115 \\
Blood Donation   & 0.567 & 0.582 & 0.593 & 0.026 & 0.663 & 0.530 & -0.133 \\
\midrule
\emph{Average}    & 0.680 & 0.664 & 0.707 & 0.057 & 0.714 & 0.662 & -0.052 \\
\bottomrule
\end{tabularx}
\caption{Ablation study on the impact of feature synonyms and contextual preambles on LoID's performance. Left: Effect of feature description variants. Performance varies only modestly across paraphrases, indicating that LoID is robust to changes in feature descriptions. Right: Effect of contextual preambles. Adding contextual information generally degrades performance, potentially because the preamble biases the LLM toward its task-oriented generations rather than its underlying beliefs.}
\label{tab:adon}
\end{table*}
\begin{table}[t]
\centering

\small
\begin{tabular}{p{3.5cm} p{10cm}}
\toprule
\textbf{Condition} & \textbf{Example Prompt} \\
\midrule
\multirow{3}{3.5cm}{\textbf{No Preamble (Baseline)}} & Consider a person who is older. Is this person at risk of stroke? Answer: \\
 & \\
 & Consider a person who has donated blood more recently. Does this person have blood donation likelihood? Answer: \\
 & \\
 & Consider a person who has higher ejection fraction. Does this person have poor outcome? Answer: \\
\midrule
\multirow{3}{3.5cm}{\textbf{With Preamble}} & \textbf{Stroke risk increases significantly with age due to arterial stiffening and accumulated vascular damage. Based on this medical knowledge,} consider a person who is older. Is this person at risk of stroke? Answer: \\
 & \\
 & \textbf{People who have donated blood recently are typically more engaged and committed donors. Based on this behavioral pattern,} consider a person who has donated blood more recently. Does this person have blood donation likelihood? Answer: \\
 & \\
 & \textbf{Higher ejection fraction indicates better cardiac pump function and improved prognosis. Based on this cardiac physiology,} consider a person who has higher ejection fraction. Does this person have poor outcome? Answer: \\
\bottomrule
\end{tabular}
\caption{Example prompts comparing baseline (no preamble) with contextual preamble condition.}
\label{tab:preamble_examples}
\end{table}

\paragraph{Results} Contextual preambles degrade performance on average, reducing mean AUC by 5.2 points (0.714 $\rightarrow$ 0.662). The effect is inconsistent: Adult Income (+2.6 points) and Stroke (+1.3 points) benefit modestly from preambles, whereas Heart Failure (-11.5 points) and Blood Donation (-13.3 points) experience substantial degradation. This suggests that contextual reasoning can shift the model toward task-oriented generation rather than preserving the underlying token probability distributions from which LoID extracts priors. A similar limitation applies to generation-based approaches such as AutoElicit, where the model is explicitly prompted to generate its beliefs. In contrast, LoID extracts priors directly from token-level probabilities without relying on generated responses. These results therefore suggest that minimal prompting is preferable for logit-based prior extraction, as additional contextual framing may inadvertently steer the model toward generation rather than exposing its underlying probability mass.

\subsection{Datasets and OOD splits}
\label{subsec:datasets}

\subsubsection{Public datasets}

\textbf{Adult Income}, \cite{adult_2}: Predicts whether an individual's annual income exceeds \$50K using demographic and educational features such as age, work class, education, marital status, occupation, race, sex, and hours worked per week. The binary classification task is based on 48,842 records from the U.S. Census Bureau Adult dataset.  

\textbf{Cervical Cancer}, \cite{cervical_cancer}: Predicts the risk of cervical cancer using demographic factors, sexual health history, and medical test results. Features include age, number of sexual partners, age at first intercourse, smoking habits, hormonal contraceptive use, and HPV test results. The dataset contains 858 patient records collected from clinics in Colombia.  

\textbf{Chronic Kidney Disease} \cite{chronic_kidney_disease_336}: Classifies the presence of chronic kidney disease using clinical and laboratory measurements, including blood pressure, glucose level, blood urea, serum creatinine, hemoglobin, and albumin. The binary classification task is based on 400 patient records from multiple hospitals.  

\textbf{Bank Marketing}, \cite{bank_marketing_222}: Predicts whether a client will subscribe to a term deposit based on demographic attributes and call-related features such as age, job type, marital status, education, balance, contact communication type, and duration of the last call. The dataset contains 45,211 records collected from Portuguese banking campaigns.  

\textbf{Blood Transfusion Service Center}, \cite{blood_transfusion_service_center_176}: Predicts whether a donor will donate blood again using historical donation behavior, including recency, frequency, monetary amount of previous donations, and time since first donation. The dataset consists of 748 donor records from a blood transfusion center in Taiwan.  

\textbf{Stroke}, \cite{fedesoriano2020stroke}: Predicts the risk of stroke using demographic and health-related features such as age, hypertension, heart disease, glucose level, BMI, smoking status, and gender. The binary classification task is based on 5110 patient records collected from hospitals in India.  

\textbf{Heart Failure}, \cite{heart_failure_clinical_records_519}: Predicts mortality due to heart failure using clinical measurements including age, ejection fraction, serum creatinine, serum sodium, platelets, and smoking status. The dataset contains 299 patient records collected from hospitals in Portugal.  

\textbf{Diabetes Prediction}, \cite{mustafatz2023diabetes}: Classifies diabetes risk using blood glucose levels, BMI, and other demographic and health-related features. The dataset contains 520 patient records and is used to predict the onset of diabetes in individuals.  

\textbf{Give Me Some Credit}, \cite{givemesomecredit2011}: Predicts the probability of financial distress within two years using demographic and credit-related features such as age, monthly income, number of past delinquencies, number of open credit lines, and debt-to-income ratio. The dataset consists of 150,000 records from U.S. financial institutions.  

\textbf{Airfoil Self-Noise}, \cite{airfoil_self-noise_291}: Predicts the noise levels generated by an airfoil using measurements collected by NASA. Features include frequency, angle of attack, chord length, free-stream velocity, and suction side displacement thickness. The regression task is based on 1,503 measurements of airfoil noise under different operating conditions.  

\textbf{Bike Sharing}, \cite{bike_sharing_275}: Predicts hourly bicycle rental demand based on temporal and weather-related features, including hour of the day, day of the week, temperature, humidity, wind speed, and season. The dataset consists of 17,379 hourly records collected from a bike-sharing system in Washington, D.C.  

\textbf{Combined Cycle Power Plant}, \cite{combined_cycle_power_plant_294}: Predicts net hourly electrical energy output using ambient variables such as temperature, pressure, humidity, and exhaust vacuum. The regression task contains 9,568 records collected from a combined cycle power plant over six years.  

\textbf{Concrete Compressive Strength}, \cite{concrete_compressive_strength_165}: Predicts the compressive strength of concrete using mixture components such as cement, blast furnace slag, fly ash, water, superplasticizer, coarse aggregate, and fine aggregate, as well as curing age. The dataset contains 1,030 records of concrete mixtures tested under controlled conditions.  

\textbf{Communities and Crime}, \cite{communities_and_crime_183}: Predicts per-capita violent crime rates in U.S. communities using demographic, socioeconomic, and law enforcement features, including population, income, employment, education, and police presence. The dataset contains 1,994 community records compiled from U.S. government sources.  

\subsubsection{Private datasets}
\paragraph{Head and Neck Cancer Recurrence Dataset}
In collaboration with Henry Ford Health, we compiled a binary classification dataset for Head and Neck Cancer Recurrence. This dataset includes comprehensive features about patient demographics, tumor characteristics, treatment regimens, and clinical staging, and predicts the likelihood of cancer recurrence following radiotherapy. The dataset encompasses 508 patients with head and neck cancers, capturing clinical variables including comorbidity indices (Charlson score), smoking history (pack-years), HPV status, primary tumor site localization, and detailed radiation therapy parameters such as treatment duration, prescription dose, and the critical time interval between biopsy and treatment initiation. 

This dataset presents a significant class imbalance (390 non-recurrence vs. 118 recurrence cases), reflecting real-world clinical distributions. The dataset is particularly valuable for evaluating model performance under distributional shift, as we created an out-of-distribution train set focusing on elderly patients ($age \geq 75$), representing a clinically meaningful subpopulation that is not present in any LLM pre-training corpora. 

Moreover, this dataset highlights why our approach is necessary. Due to its limited sample size, some patterns learned from the data are not medically plausible. Table~\ref{tab:HNC} shows several noticeable relationships between features and the target variable. For instance, the data suggests that marijuana use is associated with a lower risk of Head and Neck Cancer recurrence, which contradicts established medical evidence \cite{kim2020cannabis_hnc}. This discrepancy is likely due to sampling bias or confounding factors in the small dataset.

LoID priors address this issue by incorporating domain knowledge into the model. In this case, they can enforce a prior that reflects a more realistic relationship between marijuana use and cancer recurrence. As a result, the model is guided away from spurious correlations, making it more reliable and better suited for generalization to real-world populations.

\begin{table}[h]
\centering
\small
\renewcommand{\arraystretch}{1.2} 
\setlength{\extrarowheight}{2pt} 
\setlength{\tabcolsep}{6pt} 

\begin{tabular}{lcc}
\hline
\textbf{Recurrence} & \textbf{Yes} & \textbf{No} \\
\hline
\textbf{Sample Size} & 118 & 390 \\
\hline
\multicolumn{3}{l}{\textbf{CONTINUOUS FEATURES (Mean [Standard Deviation])}} \\[2pt]
Pack Years & 41.3 (38.1) & 27.3 (28.7) \\
Biopsy-RT Duration & 44.6 (28.2) & 46.2 (47.5) \\
Average Income & 82348.7 (32140.4) & 90597.7 (36881.5) \\
\hline
\multicolumn{3}{l}{\textbf{CATEGORICAL FEATURES (Mode [\%])}} \\[2pt]
Clinical N & 2a (50.0) & 0 (29.0) \\
Clinical T & 2 (53.7) & 2 (57.6) \\
Radiation Type & IMRT Standard (50.4) & IMRT Vmat (52.2) \\
Alcohol & Frequent (41.3) & No Alcohol Use (45.8) \\
Smoking History & Former Use (45.4) & Never Smoker (41.5) \\
Grade of Differentiation & Moderate (68.7) & Moderate (79.0) \\
Neck RT & Bilateral (85.1) & Bilateral (91.7) \\
Tobacco & Yes (54.5) & No (63.2) \\
Marijuana & No (87.6) & No (73.5) \\
First Primary Site = Larynx-supraglottis & No (73.6) & No (82.0) \\
First Primary Site = Larynx-glottis & No (82.6) & No (92.0) \\
First Primary Site = Oropharynx-base of tongue & No (78.5) & No (69.5) \\
First Primary Site = Oropharynx-tonsil & No (79.3) & No (70.5) \\
Concurrent Chemotherapy & No (90.1) & No (95.5) \\
\hline
\end{tabular}
\caption{Notable feature--target relationships in the Head and Neck Cancer Recurrence dataset. While some associations, such as those involving tobacco use, pack years, and alcohol consumption, align with clinical expectations, others are misleading. In particular, a larger proportion of marijuana users appear in the non-recurrence group compared to the recurrence group, which may incorrectly suggest a protective effect. A standard logistic regression model may learn this spurious correlation, despite it lacking medical validity.}
\label{tab:HNC}
\end{table}

\subsection{Out-of-Distribution splits}
\label{app:ood_splits}

Our goal in this research is to evaluate how well a model trained on a small, specific population generalizes to a larger population. To simulate this, we create conditions where only a subset of the full population is available for training, but the aim remains to extract meaningful patterns from the data. Specifically, we first generate train and test sets representing the entire population using a standard 80-20 shuffle split. From the training set, we then select a subset that excludes certain samples entirely, creating a subpopulation that is not fully representative of the overall population. This setup reflects real-world scenarios, such as Head and Neck Cancer Recurrence, where the limited number of training samples can hinder pattern discovery. Table~\ref{tab:ood_splits} summarizes the split criteria for all datasets.

\textbf{Binary Classification:}
\begin{itemize}
    \item \textbf{Adult (Income Prediction)}: Training set excludes individuals age $< 28$.
    \item \textbf{Chronic Kidney Disease}: Training set excludes patients with blood urea $< 61.75$ mg/dL.
    \item \textbf{Cervical Cancer}: Training set excludes women whose age at first sexual intercourse $\leq 15$.
    \item \textbf{Blood Donation}: Training set excludes donors with recency $> 14$ months (i.e., recent donors are excluded).
    \item \textbf{Stroke}: Training set includes only patients age $\leq 32$.
    \item \textbf{Heart Failure}: Training set includes only patients with ejection fraction $\geq 30\%$.
    \item \textbf{Diabetes Prediction}: Training set includes patients with blood glucose $\leq 126$ mg/dL; testing set includes hyperglycemic patients with glucose $\geq 126$ mg/dL.
    \item \textbf{Bank Marketing}: Training set excludes calls with duration $> 127$ seconds.
    \item \textbf{Head \& Neck Cancer (HNC)}: Training set includes patients age $\leq 75$.
    \item \textbf{Good Credit Score}: Training set excludes individuals age $> 31$.
\end{itemize}

\textbf{Regression:}
\begin{itemize}
    \item \textbf{Airfoil Noise}: Training set includes low-frequency conditions (frequency $< 1600$ Hz).
    \item \textbf{Bike Sharing}: Training set includes early hours (hour $\leq 6$).
    \item \textbf{Combined Cycle Power}: Training set includes moderate temperatures (AT $\leq 13.51^\circ$C).
    \item \textbf{Communities \& Crime}: Training set includes racially diverse areas (racePctWhite $\leq 0.98$).
    \item \textbf{Concrete Strength}: Training set includes high-cement mixes ($> 350$ kg/m³), where mixture interactions change.
\end{itemize}

\begin{table}[t]
\centering

\small
\begin{tabular}{l l l r r}
\toprule
\textbf{Dataset} & \textbf{Split Feature} & \textbf{Excluded Region} & \textbf{$P_{\text{IID}}$} & \textbf{$P_{\text{OOD}}$} \\
\midrule
\multicolumn{5}{l}{\textit{Binary Classification}} \\
Adult Income & Age & Age $<$ 28 & 0.809 & 0.414 \\
Cervical Cancer & First sexual intercourse & Age $<$ 15 & 0.870 & 0.440 \\
Chronic Kidney Disease & Blood urea & BU $<$ 61.75 mg/dL & 0.993 & 0.822 \\
Diabetes Prediction & Blood glucose & Glucose $>$ 126 mg/dL & 0.963 & 0.883 \\
Stroke & Age & Age $>$ 32 & 0.842 & 0.608 \\
Heart Failure & Ejection fraction & EF $<$ 30\% & 0.849 & 0.746 \\
Bank Marketing & Call duration & Duration $>$ 127 sec & 0.878 & 0.780 \\
Blood Donation & Recency & Recency $>$ 14 months & 0.748 & 0.572 \\
Head \& Neck Cancer & Age & Age $>$ 75 & 0.998 & 0.542 \\
Good Credit Score & Age & Age $>$ 31 & 0.790 & 0.710 \\
\midrule
\multicolumn{5}{l}{\textit{Regression}} \\
Concrete Strength & Cement & Cement $>$ 350 kg/m³ & 98.1 & 469.2 \\
Communities \& Crime & Race (White \%) & White $>$ 98\% & 0.007 & 0.171 \\
Airfoil Noise & Frequency & Freq $<$ 1600 Hz & 23.7 & 77.2 \\
Combined Cycle Power & Temperature & Temp $>$ 13.51°C & 21.2 & 514.9 \\
Bike Sharing & Hour & Hour $>$ 6 & 19030 & 77858 \\
\bottomrule
\end{tabular}
\caption{Out-of-distribution split specifications for all datasets. Train/test splits are created by partitioning on a single feature to induce distributional shift. Choosing the OOD train samples from the IID train set ensures that the IID test set and OOD train set don't overlap.}
\label{tab:ood_splits}
\end{table}

\subsection{Additional baselines}
\label{subsec:addbase}

\subsubsection{Correlation analysis}
\label{sec:correlation}
In addition to the baselines mentioned in Table~\ref{tab:auc-kl-combined}, we investigate whether simple feature-target correlations can capture relationships similar to those extracted by LoID. Specifically, we compute the Pearson correlation between each feature and the target using the training data and directly use these correlation coefficients as the corresponding model coefficients. Table~\ref{tab:baselines_comparison} compares LoID against this correlation-based baseline (the ``Corr. Coeff'' column). As expected, correlations estimated from OOD train sets can be misleading. LoID outperforms the correlation baseline on most datasets. This suggests that the priors extracted by LoID capture information beyond simple feature-target correlations. Nevertheless, correlations estimated from OOD train sets can sometimes provide competitive performance. This is evident for Bank Marketing, where LoID also fails because the task exhibits dataset-specific behavior. For Combined Cycle Power, the $\mathcal{N}(0,1)$ baseline already recovers much of the performance, indicating that the OOD train set is still relatively representative of the test distribution after regularization. Consequently, the correlation-based baseline also achieves competitive performance. These results further support our finding that relationships estimated directly from the training data do not consistently capture the distributional structure needed for OOD generalization.

\subsubsection{L1 and L2 Regularization}
\label{sec:L1L2}

We compare LoID against traditional Bayesian regularization approaches to determine whether its gains arise from informative LLM-derived priors rather than regularization alone. We implement two Bayesian regularization baselines: (1) \textbf{Bayesian L2}, which places Gaussian priors on coefficients,
$\beta \sim \mathcal{N}(0,1)$ (equivalent to the $\mathcal{N}(0,1)$ column of Table \ref{tab:auc-kl-combined}), and (2) \textbf{Bayesian L1}, which places Laplace priors on coefficients,
$\beta \sim \mathrm{Laplace}(0,1)$.
Both approaches use zero-centered priors and are trained using NUTS with 1,000 tuning iterations followed by 1,000 posterior samples and a target acceptance rate of 0.95.

Table~\ref{tab:baselines_comparison} summarizes results across all datasets in our evaluation. Overall, LoID outperforms both Bayesian L1 and Bayesian L2 on most OOD classification tasks, with particularly large gains on medical datasets. For example, on Kidney Disease, LoID achieves an AUC of 0.901 compared to 0.715 and 0.706 for Bayesian L2 and L1, respectively. Similarly, on Stroke Prediction, LoID reaches 0.694 AUC compared to 0.618 and 0.633 for the two regularization baselines. Large improvements are also observed on Cervical Cancer (0.681 vs. 0.448 and 0.462) and Blood Donation (0.663 vs. 0.572 and 0.579).

These results suggest that the primary benefit of LoID comes from informative prior means rather than coefficient shrinkage. Both Bayesian L1 and L2 regularization constrain coefficient magnitudes toward zero but cannot encode directional beliefs about feature effects. The relatively small differences between Bayesian L1 and L2 across most datasets further indicate that the choice of regularization family is less important than the informativeness of the prior itself.

For regression tasks, the pattern is more mixed. LoID achieves the lowest MSE on Concrete Strength, Communities and Crime, and Airfoil Noise, while Bayesian L2 performs substantially better on Combined Cycle Power and slightly better on Bike Sharing. Section \ref{sec:fails} explains the reason behind these failures in depth. 

Overall, the average gap closed (42\% for LoID vs. 18\% for L2 and 20\% for L1 regularization) indicates that LoID's improvements stem from the injection of transferable domain knowledge through LLM-derived priors, rather than from regularization alone.

\begin{table}[t]
\centering
\small

\begin{tabular*}{\linewidth}{@{\extracolsep{\fill}}lcccc@{}}
\toprule
\textbf{Dataset} &
\textbf{LoID} &
\textbf{L2: $\mathcal{N}(0,1)$} &
\textbf{L1: $\mathrm{Laplace}(0,1)$} &
\textbf{Corr. Coeff} \\
\midrule

\multicolumn{5}{l}{\textit{Binary Classification (Test AUC)}} \\
Head \& Neck Cancer & \textbf{0.696} & 0.567 & 0.575 & 0.566 \\
Adult Income & \textbf{0.745} & 0.510 & 0.510 & 0.529 \\
Cervical Cancer & \textbf{0.681} & 0.448 & 0.462 & 0.450 \\
Kidney Disease & \textbf{0.901} & 0.715 & 0.706 & 0.575 \\
Diabetes & \textbf{0.906} & 0.853 & 0.843 & 0.873 \\
Stroke & \textbf{0.694} & 0.618 & 0.633 & 0.554 \\
Heart Failure & \textbf{0.753} & 0.742 & 0.751 & 0.742 \\
Bank Marketing & 0.652 & 0.780 & \textbf{0.782} & 0.770 \\
Blood Donation & \textbf{0.663} & 0.572 & 0.579 & 0.574 \\
Good Credit Score & \textbf{0.760} & 0.710 & 0.716 & 0.376 \\
\midrule

\multicolumn{5}{l}{\textit{Regression (Test MSE)}} \\
Concrete Strength & \textbf{120.59} & 165.72 & 172.20 & 2292.2 \\
Communities \& Crime & \textbf{0.07} & 0.17 & 0.17 & 0.18 \\
Airfoil Noise & \textbf{40.51} & 51.11 & 51.47 & 51.4 \\
Combined Cycle Power & 509.92 & \textbf{23.55} & 24.03 & 60.1 \\
Bike Sharing & 78351.00 & 71890.00 & 72197.00 & \textbf{69969} \\
\midrule

\emph{Average gap closed} & \textbf{42\%} & 18\% & 20\% & 13\% \\
\bottomrule
\end{tabular*}

\caption{Comparison of LoID against Bayesian regularization baselines and correlation-based priors on OOD test sets. Bayesian L2 uses $\mathcal{N}(0,1)$ priors with no tuning, while Bayesian L1 uses $\mathrm{Laplace}(0,1)$ priors. Correlation-based priors use feature correlations estimated from the training data. Best results are shown in bold.}
\label{tab:baselines_comparison}
\end{table}

\subsection{Computational efficiency}
\label{sec:comp_eff}
We compare the computational complexity and runtime of LoID with AutoElicit and LLMProcesses.
For a dataset with $d$ features, the LoID method queries the LLM using
$N_{\text{sent}}$ sentence formulations per feature and extracts the probabilities
of two tokens (\textit{positive} and \textit{negative}). Its time complexity is:
\[
T_{\text{logit}} = \mathcal{O}(d \times N_{\text{sent}} \times T_{\text{LLM}}),
\]
where $T_{\text{LLM}}$ denotes the inference time of a single LLM query.
In our experiments, $N_{\text{sent}} = 10$, and both token probabilities (\textit{positive} and \textit{negative}) are read from a single forward pass per prompt, resulting in 10 LLM queries per feature.

AutoElicit employs an interactive multi-round elicitation process with iterative refinement,
leading to a higher complexity:
\[
T_{\text{AE}} = \mathcal{O}(d \times N_{\text{rounds}} \times N_{\text{queries}} \times T_{\text{LLM}}),
\]
where $N_{\text{rounds}} = 10 $ and $N_{\text{queries}} = 10$.
This results in more than 100 LLM queries per feature on average.

Empirically, this yields a consistent \textbf{10$\times$ speedup} in prior elicitation
for LoID across all evaluated datasets.

LLMProcesses follows a fundamentally different paradigm: it generates predictions for each test
instance by sampling from the LLM $N_{\text{samples}}$ times using in-context examples.
For a test set of size $n_{\text{test}}$ with $n_{\text{train}}$ training examples in context,
its complexity is:
\[
T_{\text{LLMP}} =
\mathcal{O}(n_{\text{test}} \times N_{\text{samples}} \times T_{\text{LLM}}(n_{\text{train}} \cdot d)),
\]
where $T_{\text{LLM}}(n_{\text{train}} \cdot d)$ reflects the increased inference time
caused by long prompts containing $n_{\text{train}}$ in-context examples.
Unlike LoID and AutoElicit, which perform prior elicitation once per dataset,
LLMProcesses scales linearly with the test set size and incurs substantially longer runtimes.


\subsection{Resources and reproducibility}

\paragraph{Code availability} To ensure reproducibility, we will release our full codebase, including detailed instructions for running the experiments and reproducing all reported results, including the scripts used to generate all figures and tables in the paper.

\paragraph{Data availability} All datasets used in this study except Head and Neck Cancer Recurrence are publicly available through open-access repositories. Further details and references are provided in Section \ref{subsec:datasets}. All preprocessing steps required to obtain the final train/test splits used in our experiments, along with the preprocessed data, are available in our code release.

The Head and Neck Cancer Recurrence dataset will be released upon publication.

\paragraph{Results variability} 
To assess the stability and determinism of our approach, all experiments were repeated five times. The resulting 95\% confidence intervals for AUC were effectively zero (0.000), indicating that our method yields highly consistent results across runs.

\paragraph{Computational resources}
\begin{enumerate}
\item \textbf{Infrastructure:} Experiments were run on a workstation equipped with \textbf{2 NVIDIA RTX 6000 Ada Generation GPUs} (each with 50GB VRAM), a \textbf{Threadripper PRO 7985WX CPU} with \textbf{64 cores (128 threads)} at up to \textbf{5.37 GHz}, and \textbf{256 GB of RAM}.

\item \textbf{Model inference:} We used \textbf{Gemma-2-27B} as our main LLM. All inference was performed using both GPUs in parallel. Each feature prompt required about \textbf{50--70 seconds} to complete across 10 paraphrases, leading to approximately \textbf{60 seconds * \#features} per dataset.

\item \textbf{Preprocessing:} Minimal preprocessing was required and was handled efficiently on a single CPU.

\item \textbf{Additional resources:} Exploratory analyses, including feature selection, were performed using \textbf{Claude-3.5-Sonnet via OpenRouter}. Each call took \textbf{2--3 seconds} and used one CPU thread.

\end{enumerate}

\end{document}